\documentclass{elsart5p}

\usepackage{natbib}
\usepackage{graphicx}
\usepackage{amssymb}
\usepackage{amsmath}
\begin{document}
\begin{frontmatter}



\title{Dynamic Muscle Fatigue Evaluation in Virtual Working Environment}


\author[ECN]{Liang MA\corauthref{cor}}
\ead{liang.ma@irccyn.ec-nantes.fr}
\author[ECN]{Damien CHABLAT}
\author[ECN]{Fouad BENNIS}
\author[Tsinghua]{Wei ZHANG}
\address[ECN]{Institut de Recherche en Communications et en Cybern\'etique de Nantes, UMR CNRS 6597, \\\'Ecole Centrale de Nantes\\1,rue de la No\"{e} - BP 92 101 - 44321 Nantes CEDEX 03, France}
\address[Tsinghua]{Department of Industrial Engineering, Tsinghua University\\100084, Beijing, P.R.China}
\corauth[cor]{Corresponding author: Tel:+33 2 40 37 69 58; Fax:+33 2 40 37 69 30}

\begin{abstract}

Musculoskeletal disorder (MSD) is one of the major health problems in mechanical work especially in manual handling jobs. Muscle fatigue is believed to be the main reason for MSD. Posture analysis techniques have been used to expose MSD risks of the work, but most of the conventional methods are only suitable for static posture analysis. Meanwhile the subjective influences from the inspectors can result differences in the risk assessment. Another disadvantage is that the evaluation has to be taken place in the workshop, so it is impossible to avoid some design defects before data collection in the field environment and it is time consuming. In order to enhance the efficiency of ergonomic MSD risk evaluation and avoid subjective influences, we develop a new muscle fatigue model and a new fatigue index to evaluate the human muscle fatigue during manual handling jobs in this paper. Our new fatigue model is closely related to the muscle load during working procedure so that it can be used to evaluate the dynamic working process. This muscle fatigue model is mathematically validated and it is to be further experimental validated and integrated into a virtual working environment to evaluate the muscle fatigue and predict the MSD risks quickly and objectively.

\vspace{0.3cm}
\noindent{\textbf{Relevance to industry}}
\vspace{0.4cm}

Muscle fatigue is one of the main reasons causing MSDs in industry, especially for mechanical work. Correct evaluation of muscle fatigue is necessary to determine work-rest regimens and reduce the risks of MSD.

\end{abstract}

\begin{keyword}
Muscle fatigue; muscle fatigue index; muscle fatigue model; virtual reality; virtual environment
\end{keyword}

\end{frontmatter}

\section{Introduction}

Musculoskeletal disorder is defined as injuries and disorders to muscles, nerves, tendons, ligaments, joints, cartilage and spinal discs and it does not include injuries resulting from slips, trips, falls or similar accidents \citep{Oregon2000}. From the report of HSE \citep{HSE2005} and 1993 WA State Fund compensable claims \citep{Ergweb}, over 50\% of workers  in industry have suffered from musculoskeletal disorders. According to the analysis in the book \textit{Occupational Biomechanics} \citep{Chaffin1999}, overexertion of muscle force or frequent high muscle load is the main reason to cause muscle fatigue, and further more, it results in acute muscle fatigue, pain in muscles and the worst functional disability in muscles and other tissues of human body. Hence, it becomes an important mission for ergonomists to find an efficient method to evaluate the muscle fatigue and further more to decrease MSD risks caused by muscle fatigue.

There are various tools available for ergonomists to evaluate the MSD risks, most of them are listed and compared in paper \citep{GUANGYANLI1999}. For general posture analysis, Posturegram, Ovako Working Posture Analyzing System (OWAS), posture targeting and Quick Exposure Check (QEC) were developed. Further more, some special tools are designed for specified parts of human body. For example, Rapid Upper Limb Assessment (RULA) is designed for assessing the severity of postural loading and is particularly applicable for sedentary jobs. The similar systems include HAMA, PLIBEL and so on \citep{Neville2004}. 

In these methods, several ``risk factors'' of mechanical jobs are taken into consideration, like  physical work load factors, psychosocial stressors and individual factors. The level of exposure to physical work load can be evaluated with respect to intensity (or magnitude), repetitiveness and duration \citep{GUANGYANLI1999}. Using these tools, MSD risks can be effectively reduced, but there are still several limitations. At first, even just for lifting job, five prevailing tools (NIOSH lifting index, ACGIH TLV, 3DSSPP, WA L \& I, Snook) were developed. The evaluation results of them for a same task were different, and sometimes even contradictory \citep{Steven2007}. Second, most of the traditional methods have to be carried out in field environment, there is no immediate result from the observation and it is time consuming for the repetition. Further more, there are different evaluation results from different subjects even using a same method \citep{Dan2007}. At last,only posture information and limited working conditions are considered in these methods, and thus they are not able to estimate the MSD risks into detailed analysis and can not further enhance the safety of the work. The most significant problem is that muscle fatigue prediction and assessment tool is still in blank in conventional methods.

In order to evaluate the human work objectively and quickly, virtual human techniques have been developed to facilitate the ergonomic evaluation. For example, Jack \citep{BADLER1999}, ErgoMan \citep{Shaub1997}, 3DSSPP \citep{Chaffin1969}, Santos \citep{VSR2004,Vignes2004} and so on, these human modeling tools are used in field of automotive, military, aerospace and industrial engineering. These human modeling tools are mainly used for visualization to provide information about body posture, reachability and field of view and so on. In paper \citep{Uma2006}, a method to link virtual environment (Jack) and quantitative ergonomic analysis tools (RULA) in real time for occupational ergonomics is presented, and it allows that ergonomic evaluation can be carried out in real time in their prototype system. But until today, these is still no fatigue index available in these virtual human tools for dynamic working process. It is necessary to develop the muscle fatigue model and then integrate it into the virtual human softwares to evaluate the muscle fatigue and analyze the mechanical work into details.

For objectively predicting muscle fatigue, several muscle fatigue models and fatigue index have been proposed in publications. In a series of publications  \citep{Wexler1997,Wexler20001,Wexler20002,Wexler20022,Wexler20021,Wexler2003}, Wexler and his colleges have proposed a new muscle fatigue model based on $Ca^{2+}$ cross-bridge mechanism and verified the model with stimulation experiments, but it is mainly based on physiological mechanism and it is too complex for ergonomic application. Further more, there are only parameters available for quadriceps. This model was integrated into VSR system for several muscle fatigue simulations with lots of limitations \citep{Vignes2004}. Taku Komura et al. \citep{KOMURA1999,KOMURA2000} have employed a muscle fatigue model based on force-pH relationship \citep{Giat1993} in computer graphics to visualize the muscle fatigue, but it just evaluated the muscle fatigue at a time instant and cannot evaluate the overall muscle fatigue of the working processes. Meanwhile in this pH muscle fatigue model, although the force generation capacity can be mathematically analyzed, the influences from the muscle forces are not considered enough. Rodr\'{i}guez proposes a half-joint fatigue model  \citep{Rodriguez2002,Rodriguez20032,Rodriguez20031}, more exactly a fatigue index, based on mechanical properties of muscle groups. The fatigue is quantified by accumulation of the proportion of actual holding time over the predicted maximum holding time. With this half-joint model, it can adjust human posture during a working process dynamically, but it cannot predict individual muscle fatigue due to its half-joint principle. The maximum endurance equation of this model was from static posture analysis and it is mainly suitable for evaluating static postures. In \citep{Jing2002}, a dynamic muscle model is proposed based on motor units principles, but there are just parameters available under maximum voluntary contraction situation which is rare in manual handling work.

In this paper, we are going to propose a dynamic muscle fatigue model and a fatigue index with consideration of muscle load history and personal factors. This fatigue model is going to be verified with comparison of the previous static endurance time models and several dynamic muscle fatigue models. At last, we are going to propose the experimental validation procedure in a virtual environment framework. 

\section{Dynamic Muscle Fatigue Model}
\label{sec:dynamicModel}

We believe that the fatigue of muscle is closely related to the external load of the muscle with the time and the strength of the muscle. These factors can represent the physical risk factors mentioned before: the external load the muscle with the time can include the intensity (or magnitude), repetitiveness and duration; and the muscle strength can be determined individually. Thus, the muscle force history and maximum voluntary contraction (MVC) are taken to construct our muscle fatigue model. MVC is defined as ``the force generated with feedback and encouragement, when the subject believes it is a maximal effort'' \citep{Vollestad1997}. The effect of maximum voluntary contraction on endurance time is often used in ergonomic applications to define the worker capabilities \citep{Garg2002}. Here, we are going to use MVC to describe the maximum force generation capacity of an individual muscle.

Our fatigue index is trying to describe the human feelings (subjective evaluation) about the fatigue and it is expected to have a close correlation between subjective evaluation and objective evaluation about muscle fatigue during mechanical work.

For ergonomic evaluation, there are several self-report methods to assess the physical load, body discomfort or work stress. These subjective assessments of body strain and discomfort have been the most frequently used form due to the ease of use and apparent face validity. But subjective ratings are prone to many influences. This kind of approach has too low validity and reliability. But ergonomists have to concentrate themselves on the feeling of the workers. Several authors even insist that ``If the person tells you that he is loaded and effortful, then he is loaded and effortful whatever the behavioral and performances measures may show'' \citep{GUANGYANLI1999}. Thus, to combine the subjective evaluation and objective evaluation can reduce the contradictory problems between them and can rate the muscle fatigue correctly.

The general feeling of human body about fatigue is: the larger the force is, the faster people can feel fatigue; the longer the force maintains, the more fatigue; the smaller capacity of the muscle is, the more easily we can feel fatigue. Based on this description, the equation \ref{eq:FatigueIndexDiff} is used to describe the subjective evaluation. The parameters used in the equations are listed and described in table \ref{tab:Parameters}. This equation can be explained as follows:
\begin{enumerate}
	\item $F_{cem}$ describes the capacity of the muscle after some contraction, the rest force generation ability.
	\item $F_{load}(t)/F_{cem}(t)$ is relative load which describes the muscle force relative to the capacity of the muscle at a time instant $t$. 
	\item $MVC/F_{cem}(t)$ describes the smaller capacity of the muscle, the faster the muscle get fatigued.
\end{enumerate}

\begin{table}[htbp]
	\centering
	\caption{Parameters in Dynamic Fatigue Model}
	\label{tab:Parameters}
		\begin{tabular}{lcp{0.35\textwidth}}
		\hline
		Item & Unit & Description\\
		\hline
		$MVC$					& $N$ &	Maximum voluntary contraction, maximum capacity of muscle\\
		$F_{cem}(t)$ 	& $N$ & Current exertable maximum force, current capacity of muscle\\
		$F_{load}(t)$	& $N$ & External load of muscle, the force which the muscle needs to generate\\
		$k$						& $min^{-1}$ & Constant value, 1\\
		$U$						& $min$ & Fatigue Index\\
		$\%MVC$				&				&Percentage of the voluntary maximum contraction\\
		$f_{MVC}$			&				&$\%MVC/100$\\
		\hline			
		\end{tabular}
\end{table}

\begin{equation}
\label{eq:FatigueIndexDiff}
			\frac{dU(t)}{dt} = \frac{MVC}{F_{cem}(t)} \, \frac{F_{load}(t)}{F_{cem}(t)}
\end{equation}

Meanwhile, the current maximum exertable force $F_{cem}$ is changing with the time due to the external muscle load. It makes sense that: The larger the external load, the faster $F_{cem}$ decreases; the smaller $F_{cem}$ is, the more slowly $F_{cem}$ decreases. The differential equation for $F_{cem}$ is equation \ref{eq:FcemDiff}.
		
\begin{equation}
\label{eq:FcemDiff}
			\frac{dF_{cem}(t)}{dt} = -k \frac{F_{cem}(t)}{MVC}F_{load}(t)
\end{equation}

The integration result of equation \ref{eq:FcemDiff} is equation \ref{eq:FcemInt}.
\begin{equation}
\label{eq:FcemInt}
			F_{cem}(t) = MVC \, e^{\int \limits_{0}^{t} -k \dfrac{F_{load}(u)}{MVC}du}
\end{equation}

Assume that $F(t)$ is:
\begin{equation}
\label{eq:Ft}
			F(t) = \int \limits_{0}^{t} \dfrac{F_{load}(u)}{MVC}du
\end{equation} 	

$MVC$ is a constant value for an individual person, so we can change the equation \ref{eq:FcemInt} into equation \ref{eq:FcemIntSimple}. If $F_{load}/MVC$ is constant and equals to $C $, then $F(t)=Ct$, equation \ref{eq:FcemInt} can be further simplified. This constant case can occur during static posture and static load.

\begin{equation}
\label{eq:FcemIntSimple}
			\dfrac{F_{cem}(t)}{MVC} = e^{ -kF(t)} = e^{ -kCt} 
\end{equation}

The feeling of our fatigue is a function below and which is closed related to $MVC$ and $F_{load}(t)$. $MVC$ can represent the personal factors \citep{Chaffin1999}, and $F_{load}(t)$ is the force exerted on the muscle along the time and it reflects the influences of external loads equation \ref{eq:FatigueIndexInt}. 

\begin{equation}
\label{eq:FatigueIndexInt}
		U(t) = \frac{1}{2k}e^{2kF(t)} - \frac{1}{2k}e^{2kF(0)}
\end{equation}

In this model, personal factors and external load history are considered to evaluate the muscle fatigue. It can be easily used and integrated into simulation software for real time evaluation especially for dynamic working processes. This model still needs to be mathematically validated and ergonomic experimental validated, and meanwhile muscle recovery procedure should also be included to make the fatigue index completed.

\section{Static Validation}

\subsection{Validation result}

Our dynamic muscle fatigue model is simply based on the hypothesis on the reduction of the maximum exertable force of muscle. It should be able to describe the most special case - static situations. In static posture analysis, there is no model to describe the reduction of the muscle capacity related to muscle force, but there are several models about maximum endurance time (MET) which is a measurement related to static muscular work. MET represents the maximum time during which a static load can be maintained \citep{Khalid2006}. The MET is most often calculated in relation to the percentage of the voluntary maximum contraction ($\%MVC$) or to the relative force ($f_{MVC} = \%MVC/100$) required by the task. These models which are cited from \citep{Khalid2006} are listed in table \ref{tab:StaticValidation}. 

In our dynamic model, suppose that $F_{load}(t)$ is constant, and it represents the static situation. MET is the duration in which $F_{cem}$ falls down until the current $F_{load}$. Thus, MET can be figured out in equation \ref{eq:DynamicEndurance1} and \ref{eq:DynamicEndurance2}.

\begin{equation}
\label{eq:DynamicEndurance1}
	F_{cem}(t) = MVC \, e^{\int \limits_{0}^{t} -k \dfrac{F_{load}(u)}{MVC}du} = F_{load}(t)
\end{equation}

\begin{equation}
\label{eq:DynamicEndurance2}
			t = MET = -\dfrac{ ln{\dfrac{F_{load}(t)}{MVC}}}{k\dfrac{F_{load}(t)}{MVC}} =-\dfrac{ ln(f_{MVC})}{k\, f_{MVC}}
\end{equation}

\begin{figure}[htbp]
	\centering
		\includegraphics[width=0.45\textwidth]{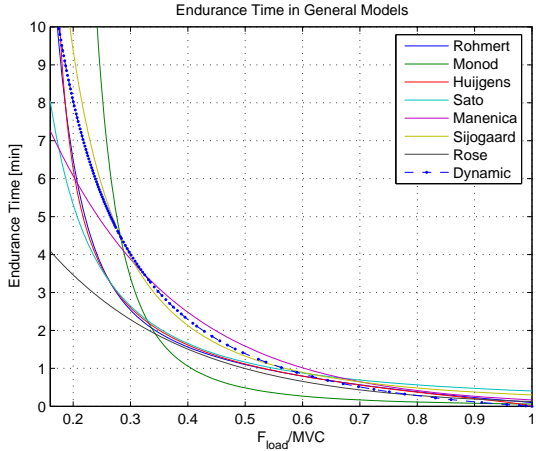}
	\caption{Endurance time in general models}
	\label{fig:general}
\end{figure}

\begin{figure}[htbp]
	\centering
		\includegraphics[width=0.45\textwidth]{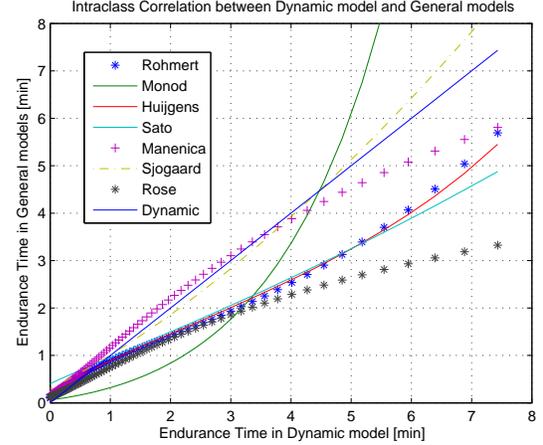}
	\caption{$ICC$ of general models}
	\label{fig:iccgeneral}\end{figure}

\begin{figure}[htbp]
	\centering
		\includegraphics[width=0.45\textwidth]{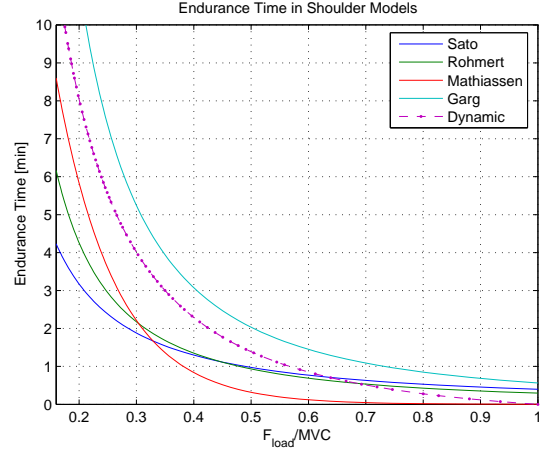}
	\caption{Endurance time in shoulder endurance models}
	\label{fig:shoulder}
\end{figure}

\begin{figure}[htbp]
	\centering
		\includegraphics[width=0.45\textwidth]{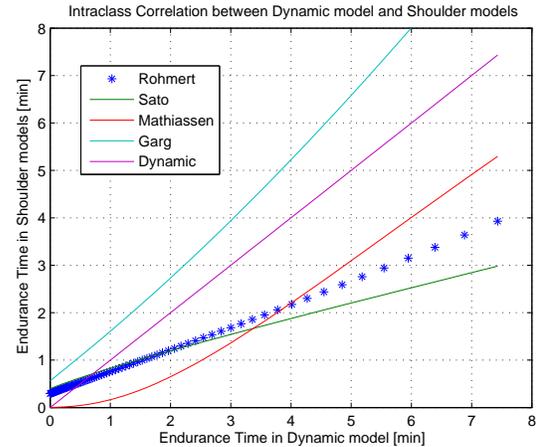}
	\caption{$ICC$ of shoulder endurance models}
	\label{fig:iccshoulder}
\end{figure}

\begin{figure}[htbp]
	\centering
		\includegraphics[width=0.45\textwidth]{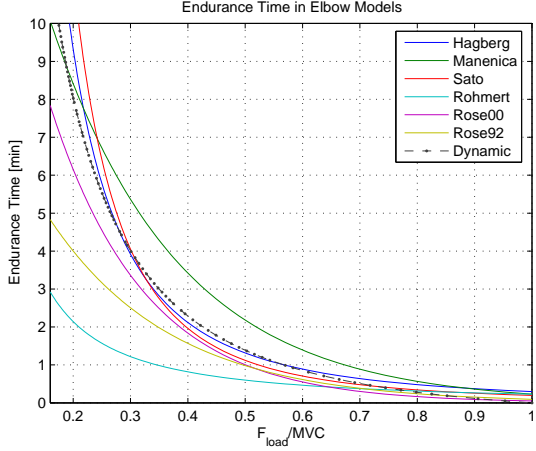}
	\caption{Endurance time in elbow endurance models}
	\label{fig:elbow}
\end{figure}

\begin{figure}[htbp]
	\centering
		\includegraphics[width=0.45\textwidth]{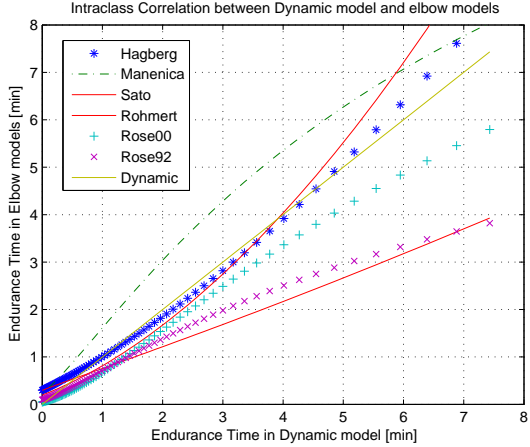}
	\caption{$ICC$ of elbow endurance models}
	\label{fig:iccelbow}
\end{figure}

\begin{figure}[htbp]
	\centering
		\includegraphics[width=0.45\textwidth]{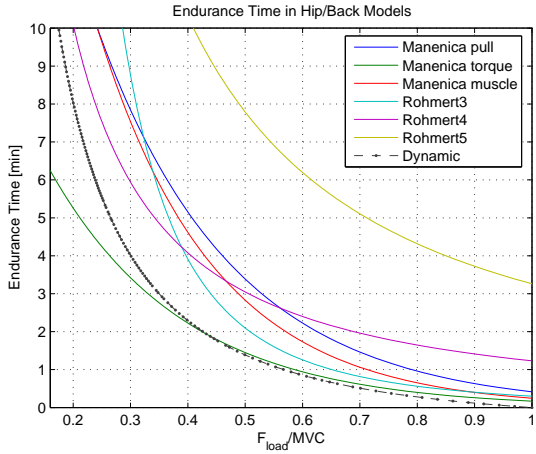}
	\caption{Endurance time in hip and back models}
	\label{fig:hip}
\end{figure}

\begin{figure}[htbp]
	\centering
		\includegraphics[width=0.45\textwidth]{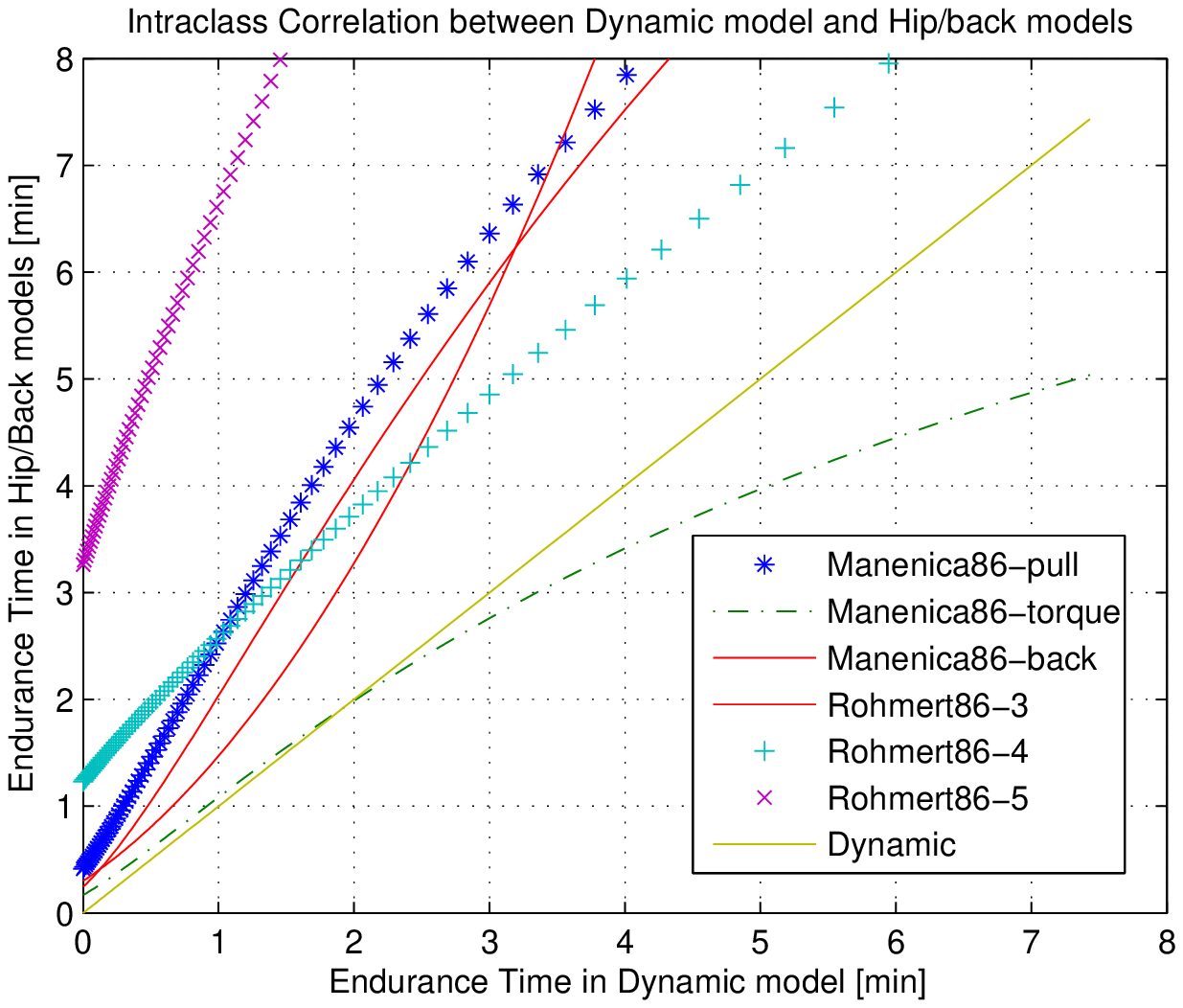}
	\caption{$ICC$ of hip/back models}
	\label{fig:icchip}
\end{figure}

In order to analyze the relationship between MET of our dynamic model and the other models, two correlation coefficients are calculated. One is Pearson's correlation $r$ in equation \ref{eq:pearson} and the other one is intraclass correlation $ICC$ in equation \ref{eq:ICC}. $r$ indicates the linear relationship between two random variables and $ICC$ can represent the similarity between two random variables. The closer $r$ is to 1, the more the two models are linear related. The closer $ICC$ is to 1, the more similar the models are. $MS_{between}$ is the mean square between different MET values in different $f_{MVC}$ values,  $MS_{within}$ is the mean square within MET values in different models at the same $f_{MVC}$ level. $p$ is the number of models in the comparison. In our case, we compare the other models with our dynamic model one by one, $k$ equals to 2. The calculation results are shown in table \ref{tab:StaticValidation} and figure \ref{fig:general} to \ref{fig:icchip}.

\begin{equation}
\label{eq:pearson}
	r = \dfrac{\sum \limits_{n}(A_{n}-\bar{A})(B_{n}-\bar{B})}{\sqrt{\sum \limits_{n}(A_{n}-\bar{A})^2\sum \limits_{n}(B_{n}-\bar{B})^2}}
\end{equation}

\begin{equation}
\label{eq:ICC}
	ICC =\frac{MS_{between}-MS_{within}}{MS_{between}+(p-1)MS_{within}}
\end{equation}

\begin{table*}[htbp]
	\centering
	\caption{Static validation results \protect \citep{Khalid2006}}
	\label{tab:StaticValidation}
		\begin{tabular}{llcc}
		\hline
		Model & Equation & r & ICC\\
		\hline
		\textbf{\textit{General models}}\\
		\hline
		Rohmert& $MET=-1.5+\frac{2.1}{f_{MVC}}-\frac{0.6}{f_{MVC}^2} +\frac{0.1}{f_{MVC}^3}$&0.9937\; &0.8820\\
		Monod and Scherrer& $MET=0.4167\,(f_{MVC}-0.14)^{-2.4}$&0.8529\;  &0.6474\\
		Huijgens& $MET=0.865\, \left[{\frac{1-f_{MVC}}{f_{MVC}-0.15}}\right]^{-2.4}$&0.9964\; &0.8800\\
		Sato et al.& $MET=0.3802 \, (f_{MVC}-0.04)^{-1.44}$&0.9992\; &0.8512\\
		Manenica& $MET=14.88\, \exp(-4.48f_{MVC})$&0.9927\; &0.9796\\
		Sjogaard& $MET=0.2991\, f_{MVC}^{-2.14}$&0.9935\; &0.9917\\
		Rose et al.& $MET=7.96\, \exp(-4.16f_{MVC})$&0.9897\; &0.7080\\
		\hline
		\textbf{\textit{Upper limbs models}}\\
		\hline
		\textit{Shoulder}\\
		Sato et al.& $MET=0.398\, f_{MVC}^{-1.29}$&0.9997\; &0.7188\\
		Rohmert et al.& $MET=0.2955\, f_{MVC}^{-1.658}$&0.9987\; &0.5626\\
		Mathiassen and Ahsberg& $MET= 40.6092\, \exp(-9.7f_{MVC}$&0.9783\; &0.7737\\
		Garg& $MET= 0.5618 \, f_{MVC}^{-1.7551}$&0.9981\; &0.9029\\
		\hline
		\textit{Elbow}\\
		Hagberg& $MET=0.298\, f_{MVC}^{-2.14}$&0.9935\; &0.9921\\
		Manenica& $MET=20.6972\, \exp(-4.5f_{MVC})$&0.9929\; &0.9271\\
		Sato et al.& $MET=0.295\, f_{MVC}^{-2.52}$&0.9838\; &0.9712\\
		Rohmert et al.& $MET=0.2285\, f_{MVC}^{-1.391}$&0.9997\; &0.7189\\
		Rose et al.2000& $MET=20.6\, \exp(-6.04f_{MVC})$&0.9986\; &0.9594\\
		Rose et al.1992& $MET=10.23\, \exp(-4.69f_{MVC})$&0.9943\; &0.7843\\
		\hline
		\textit{Hand}\\
		Manenica& $MET=20.6972\, \exp(-4.5f_{MVC})$&0.9929\; &0.9840\\
		\hline
		\textbf{\textit{Back/hip models}}\\
		\hline
		Manenica (body pull)& $MET=27.6604\, \exp(-4.2f_{MVC})$&0.9901\; &0.6585\\
		Manenica (body torque)& $MET=12.4286\, \exp(-4.3f_{MVC})$&0.9911\; &0.9447\\
		Manenica (back muscles)& $MET=32.7859\, \exp(-4.9f_{MVC})$&0.9957\; &0.7306\\
		Rohmert (posture 3)& $MET=0.3001\, f_{MVC}^{-2.803}$&0.9745\; &0.5353\\
		Rohmert (posture 4)& $MET=1.2301\, f_{MVC}^{-1.308}$&0.9989\; &0.7041\\
		Rohmert (posture 5)& $MET=3.2613\, f_{MVC}^{-1.256}$&0.9984\; &-0.057\\
		\hline							
		\end{tabular}
\end{table*}

\subsection{Discussion}

From the comparison results of the static validation, it is obvious that MET model derived from our dynamic model has a great linear correlation with the other experimental static endurance models, and almost all the Pearson's correlation $r$ are above 0.97. 

From $ICC$ column, the similarity between our dynamic model and the other models in descend sequence is: elbow models, hand model, general model, shoulder models and hip/back model. For elbow models, the average $ICC$ is approximate 0.9, but for the back/hip models, $ICC$ varies from -0.057 to 0.9447. The explanation is: in shoulder and back/hip of human body, the anatomical structure is in a much more complex way than in the elbows and hands. In this case, during these experimental models, the measurement of $MVC$ is an overall performance of the muscle group, but not an individual muscle. Meanwhile from figure \ref{fig:hip}, the differences between the experimental models for hip/back are much greater than in the other models like in elbow models (figure \ref{fig:elbow}). It can be explained as: in different working conditions (for example, different postures), the engagement of the muscles in the task in the hip/back of human body varies  due to the complexity of the structure. Thus, the error between our model and the other models is closed related to the complexity of the structure and our model can fit most of the experimental models with a high similarity.

Overall, the static validation can prove that our dynamic model can be used to predict MET in static situations. 

\section{Dynamic Validation}

\subsection{Validation result}

Static validation results have shown that our dynamic model can be used to predict the MET for general static load and even for some specific body parts. But static procedures are still quite different from dynamic situations, thus our dynamic model need to be under examination with the other dynamic models. For this objective, we are going to verify our dynamic model through comparison with some existing muscle fatigue models, quantitatively or qualitatively.

In paper \citep{Freund2001}, a muscle fatigue model was proposed and integrated into a dynamic model of forearm. In this model, the muscle was treated like a kind of reservoir, and force production capacity $S^0$ reduces with the time that the muscle is contracted. $S^0$ varies between 0 and the upper limits of the muscle force $S^l$. In this model, the recovery and decay rates depend on $S^l-S^0$ and muscle force $S$ (equation \ref{eq:Freund}). The constants $\alpha$ and $\beta$ were obtained by fitting the solution using experimental results from static endurance time test. In this model, muscle force is taken into consideration as a factor causing muscle fatigue, and further more, muscle force production capacity $S^0$ was proposed just like in our dynamic model $F_{cem}$ to describe the capacity of the muscle after performing certain task. But in this model, the force production capacity and the muscle force are decoupled with each other which is different in our model.

\begin{equation}
\label{eq:Freund}
	\frac{dS^0}{dt}=\alpha(S^l-S^0)-\beta S
\end{equation}

\begin{figure}[htbp]
	\centering
		\includegraphics[width=0.45\textwidth]{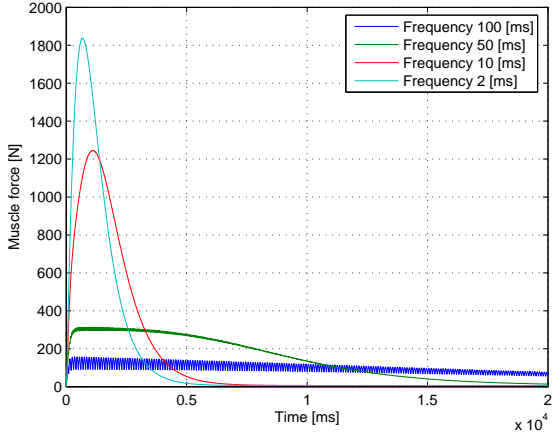}
	\caption{Maximum exertable force and time relationship in Wexler's Model}
	\label{fig:wexler1}
\end{figure}

\begin{figure}[htbp]
	\centering
		\includegraphics[width=0.45\textwidth]{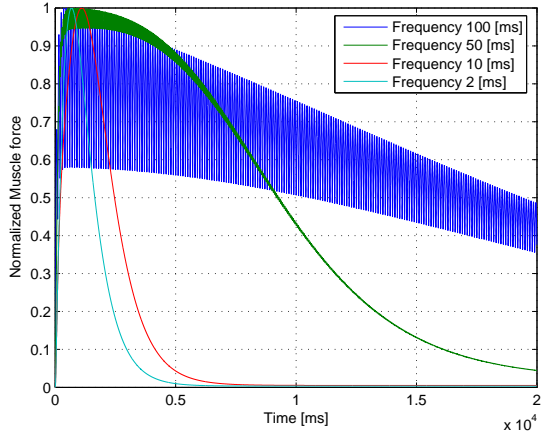}
	\caption{Normalized maximum exertable force and time relationship in Wexler's Model}
	\label{fig:wexler}
\end{figure}

Wexler's dynamic muscle fatigue model based on $Ca^{2+}$ cross-bridge mechanism can also verify our dynamic model quantitatively. This model can be used to predict the muscle force fatigue under different stimulation frequencies. From figure \ref{fig:wexler1}, it is clear that the faster the stimulation frequency is, the larger the force can be generated by the muscle. If each curve is normalized into figure \ref{fig:wexler}, it is obvious that: the larger the peak force is, the faster the muscle gets fatigue. But this model can only be used to predict the muscle force under a stimulation pattern and it can not predict or evaluate the muscle force generation capacity under a stimulation pattern when the muscle is getting fatigue. 

In the paper \citep{Jing2002}, another dynamic model of muscle activation, fatigue and recovery was presented. This model is based on biophysical mechanisms: a muscle consists of many motor units which can generate force or movement. The number of the motor units depends on the size and function of the muscle. The generated force is proportional to the activated motor units in the muscle. The brain effort $B$, fatigue property $F$ and recovery property $R$ of muscle can decide the number of activated motor units. The relationship is expressed by equation \ref{eq:activeMotor0}. The parameters in this equation is explained in table \ref{tab:ParametersInActiveMotorModel}.

\begin{figure}[htbp]
	\centering
		\includegraphics[width=0.35\textwidth]{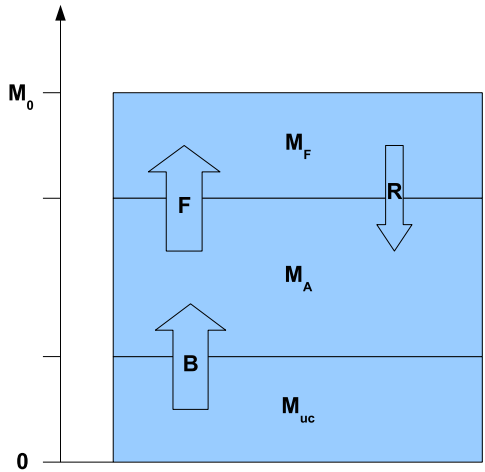}
	\caption{Illustration of the active motor model \protect \citep{Jing2002}}
	\label{fig:activemotor1}
\end{figure}

\begin{equation}
\label{eq:activeMotor0}
	\begin{array}{ccl}
	\dfrac{dM_A}{dt}&=&B\, M_{uc}-F\, M_{A}+ R\, M_{F}\\
	\dfrac{dM_F}{dt}&=&F\, M_{A} - R\, M_{F}\\
	M_{uc}&=&M_{0}-M_{A}-M_{F}\\
	\end{array}	
\end{equation}

When $t=0$ under the initial conditions of $ M_A = 0$, $M_F = 0$, $M_{uc} = M_0$, we can have equation \ref{eq:ActiveMotor1}.

\begin{equation}
\label{eq:ActiveMotor1}
\begin{split}
\frac{M_{A}(t)}{M_0}&=\frac{\gamma}{1+\gamma}+\frac{\beta}{(1+\gamma)(\beta-1-\gamma)}e^{-(1+\gamma)Ft}\\&\quad-\frac{\beta-\gamma}{\beta-1-\gamma}e^{-\beta Ft}
\end{split}
\end{equation}

\begin{table}[htbp]
	\centering
	\caption{Parameters in Active Motor Model}
	\label{tab:ParametersInActiveMotorModel}
		\begin{tabular}{lcl}
		\hline
		Item & Unit & Description\\
		\hline
		$F$ 				& $s^{-1}$ & fatigue factor, fatigue rate of motor units\\
		$R$					& $s^{-1}$ &	recovery factor, recovery rate of motor units\\
		$B$					& $s^{-1}$ & brain effort, brain active rate of motor units\\
		$M_{0}$			& & total number of motor units in the muscle\\
		$M_{A}$			& & number of activated motor units in the muscle\\
		$M_{F}$			& & number of fatigued motor units in the muscle\\
		$M_{uc}$		& & number of motor units still in the rest\\
		$\beta$			& & $B/F$\\
		$\gamma$		&	& $R/F$\\
		\hline			
		\end{tabular}
	
\end{table}

In our fatigue model, we assume that there is no recovery during mechanical work, and the workers are trying their best to finish the work which means the brain effort is quite large. In this assumption, we set  $\gamma$=0 and $\beta\rightarrow \infty$, then the equation \ref{eq:ActiveMotor2} represents the motor units which are not fatigued in the muscle and ${M_{A}(t)}/{M_0}$ represents the muscle force capacity. We can simplify the equation \ref{eq:ActiveMotor1} to equation \ref{eq:ActiveMotor2} which does have the same form of our dynamic model equation \ref{eq:FcemIntSimple}.  

\begin{equation}
\label{eq:ActiveMotor2}
	\frac{M_{A}(t)}{M_0} = e^{-Ft}-e^{-\beta Ft} =e^{-Ft}
\end{equation}

\begin{figure}[htbp]
	\centering
		\includegraphics[width=0.45\textwidth]{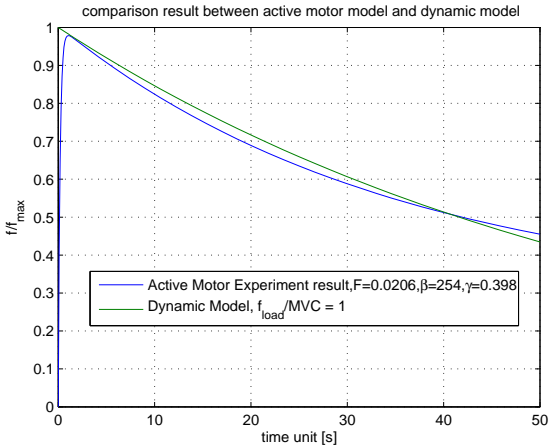}
	\caption{Comparison between the experimental result of the active motor model and dynamic model in the maximum effort}
	\label{fig:actmotor}
\end{figure}

This fatigue model has been experimentally verified \citep{Jing2002}. In the experiment, each subject performed an MVC of the right hand by gripping a handgrip device for 3 min. And the fitting curve from the experiment result has almost the same curve of our model in MVC condition (figure \ref{fig:actmotor}). In this model, $F$ and $R$ are assumed to be constant for an individual under MVC working conditions. There is no experiment result for $F$ and $R$ under the other load situations, thus this muscle fatigue model can only verify our model in MVC condition. 

\subsection{Discussion}

Through dynamic validation, our dynamic model is quantitatively and qualitatively verified with the other existing muscle fatigue models. The fatigue model used in the forearm used the same conception like in our fatigue model: the muscle force capacity is related to muscle force with time. Wexler's model based on $Ca^{2+}$ cross-bridge shows the reduction of the muscle force during the time under different stimulation frequencies, the reduction of the muscle capacity shows the same trend like in our muscle fatigue model. With comparison of the active motor model, the muscle force can be expressed in the same form under extremity situation. But in the active motor model, only parameters are available for MVC contraction case. The active motor does not supply further validation for other load situations. But we believe that the muscle load can influent the fatigue factor in the active motor model.

Overall, our dynamic model is simple and easy to use and it can evaluate the muscle fatigue during a dynamic working process.

\section{Experimental Validation}

\subsection{Objective and methodology}

After mathematical validation of our dynamic model, we are going to construct a virtual reality framework to verify the fatigue index in experimental environment. The objective of the experimental validation is to find the correlation between subjective fatigue evaluation results and objective fatigue evaluation results. In our hypothesis, if both results are highly linear related, that means the objective methods can represent the subjective feeling of fatigue and then it can be further integrated into virtual human simulation software to evaluate human muscle fatigue during manual handling work.

In order to realize our objective, two evaluation systems should be established: subjective evaluation system and objective evaluation system. Subjective evaluation methods have been mentioned in many papers, and here we are focusing on objective evaluation system. For objective evaluation, according to our model, muscle force at each time instant during the work should be calculated. It requires dynamic information like acceleration, velocity, position and external load. These information can be further input into dynamic formulas to calculate the muscle force. In order to get these information, motion capture method should be involved. As mentioned, the traditional methods have several drawbacks for ergonomic analysis, a virtual environment needs to be constructed to decrease the time cost and physical prototype cost.

\subsection{System Structure}

\begin{figure}[htbp]
	\centering
		\includegraphics[width=0.45\textwidth]{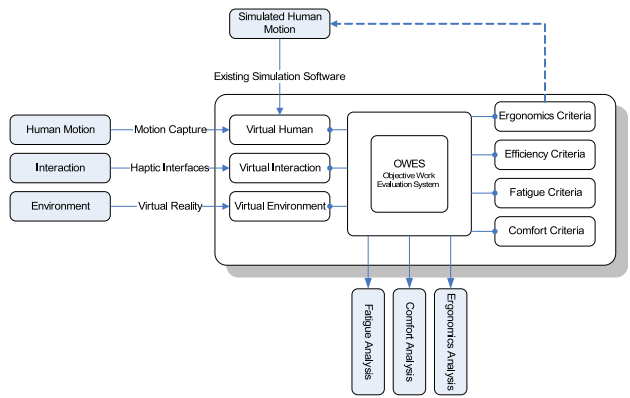}
	\caption{The framework of the dynamic work evaluation system}
	\label{fig:framework}
\end{figure}

The overall objective of the framework is to evaluate human work and predict potential human MSD risks dynamically, especially for human muscle fatigue. The function structure of the framework is shown in figure \ref{fig:framework} and discussed below. 

In order to avoid field-dependent work evaluation, virtual reality techniques and virtual human techniques are used. Immersive work simulation system should be first constructed to provide the virtual working environment. Meanwhile, virtual human should be modeled and driven by the motion capture data to map the real working procedure into the virtual environment. Haptic interfaces can be used to enable the interactions between the worker and virtual environment.

For any ergonomic analysis, data collection is the first important step. All the necessary information needs to be collected for further processing. From section \ref{sec:dynamicModel}, necessary information for dynamic manual handling jobs evaluation consists of motion, forces and personal factors. Motion capture techniques can be applied to achieve the motion information. In general, there are several kinds of tracking techniques available, like mechanical motion tracking, acoustic tracking, magnetic tracking, optical motion tracking and inertial motion tracking. Each tracking technique has its advantages and drawbacks for capturing the human motion. Hybrid motion tracking techniques can be taken to compensate the disadvantages and achieve the best motion data.

Force information can be recorded by haptic interfaces. Haptic interface is the channel via which the user can communicate with virtual objects through haptic interactions, and the interaction data between the worker and the virtual environment are also significant for evaluating other ergonomic aspects. Individual factors can be achieved from anthropometrical database and some biomechanical database.

All the information, such as motion information, force history and interaction events, is further processed into Objective Work Evaluation System (OWES). The output of the framework is evaluation results of the mechanical work. There are many ergonomic aspects of a mechanical work. For each aspect, corresponding criteria should be established to assess the dynamic work process. Further more, these criteria can also be applied just into commercial human simulation software to generate much more naturally and realistically human simulation.

Three hardware systems are employed to construct a prototype system to realize this framework: virtual simulation system, motion capture system and haptic interfaces. Virtual Simulation system consists of graphic simulation module and display module. Simulation module executes on a computer graphic station, and display module is composed of projection system and head mounted display (HMD).  Simulation module is in charge of graphic processing and display control. The projection system and HMD system can visualize the immersive environment. Motion Capture system is taken to capture the movement of the worker to collect the real working data. The Optical Motion Capture System developed by Tsinghua University VRLIB \citep{WANG2006} is used, and it can work at 25 Hz data update rate. Each key joint of human body is attached by an active marker, and the overall skeletal structure of human body can be represented by 13 markers. Meanwhile, data gloves are used to capture the movement of fingers. Haptic system is also going to be integrated into this prototype, and now we are still under preparation. 

\begin{figure}[htbp]
	\centering
		\includegraphics[width=0.45\textwidth]{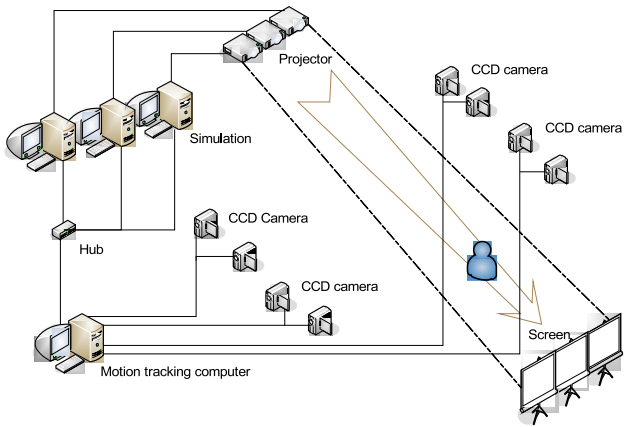}
	\caption{Scheme of Hardware system for the experimental validation}
	\label{fig:frameworkhardware}
\end{figure}

\section{Conclusion and Prospects}

In this paper, we present a new muscle fatigue model and a new fatigue index based it. The mathematical validation has proved that the new model is very simple and easy to predict the MET in static postures. The dynamic validation result proves that the new model can be further extended into dynamic working process to predict the muscle fatigue. For experimental validation, a virtual reality framework is under construction to verify the linear relation between our fatigue index and subjective evaluation result.

The prototype system cannot only evaluate the fatigue, but also the other aspects of work by extending the evaluation criteria and data collection modules. If the framework can be verified in the future, it could be integrated with the other CAD system to optimize the product design and work process design.

In future works, we will apply our results to enhance the simulation software such that they will able to produce realistic simulations.

\section*{Acknowlegement}

This research was supported by the EADS and by the R\'egion des Pays de la Loire (France) in the context of collaboration between the \'Ecole Centrale de Nantes (Nantes, France) and Tsinghua University (Beijing, P.R.China).

\bibliographystyle{elsart-harv}
\bibliography{myRef}

\begin{thebibliography}{32}
\expandafter\ifx\csname natexlab\endcsname\relax\def\natexlab#1{#1}\fi
\expandafter\ifx\csname url\endcsname\relax
  \def\url#1{\texttt{#1}}\fi
\expandafter\ifx\csname urlprefix\endcsname\relax\def\urlprefix{URL }\fi

\bibitem[{Badler et~al.(1993)Badler, Phillips, and Webber}]{BADLER1999}
Badler, N.~I., Phillips, G.~B., Webber, B.~L., 1993. Simulating Humans. Oxford
  University Press.

\bibitem[{Chaffin(1969)}]{Chaffin1969}
Chaffin, D.~B., 1969. A computerized biomechanical model-development of and use
  in studying gross body actions. Journal of Biomechanics, 429--441.

\bibitem[{Chaffin and Andrersson(1999)}]{Chaffin1999}
Chaffin, D.~B., Andrersson, G.~B., 1999. Occupational Biomechanics, 3rd
  Edition. Willey-Interscience.

\bibitem[{Ding et~al.(2000{\natexlab{a}})Ding, S.Wexler, and
  Binder-Macleod}]{Wexler20001}
Ding, J., S.Wexler, A., Binder-Macleod, S.~A., 2000{\natexlab{a}}. A predictive
  model of fatigue in human skeletal muscles. Journal of Applied Physiology,
  1322--1332.

\bibitem[{Ding et~al.(2002{\natexlab{a}})Ding, S.Wexler, and
  Binder-Macleod}]{Wexler20021}
Ding, J., S.Wexler, A., Binder-Macleod, S.~A., March 2002{\natexlab{a}}. A
  predictive fatigue model i: Predicting the effect of stimulation frequency
  and pattern on fatigue. IEEE Transactions On Neural Systems and
  Rehabilitation Engineering 10.

\bibitem[{Ding et~al.(2002{\natexlab{b}})Ding, S.Wexler, and
  Binder-Macleod}]{Wexler20022}
Ding, J., S.Wexler, A., Binder-Macleod, S.~A., March 2002{\natexlab{b}}. A
  predictive fatigue model ii: Predicting the effect of resting times on
  fatigue. IEEE Transactions On Neural Systems and Rehabilitation Engineering
  10.

\bibitem[{Ding et~al.(2000{\natexlab{b}})Ding, Wexler, and
  Binder-Macleod}]{Wexler20002}
Ding, J., Wexler, A.~S., Binder-Macleod, S.~A., 2000{\natexlab{b}}. Development
  of a mathematical model that predicts optimal muscle activation patterns by
  using brief trains. Journal of Applied Physiology, 917--925.

\bibitem[{Ding et~al.(2003)Ding, Wexler, and Binder-Macleod}]{Wexler2003}
Ding, J., Wexler, A.~S., Binder-Macleod, S.~A., 2003. Mathematical models for
  fatigue minimization during functional electrical stimulation.
  Electromyography Kinesiology, 575--588.

\bibitem[{Elahrache et~al.(2006)Elahrache, Imbeau, and Farbos}]{Khalid2006}
Elahrache, K., Imbeau, D., Farbos, B., 2006. Percentile values for determining
  maximum endurance times for static muscular work. Industrial Ergonomics,
  99--108.

\bibitem[{Freund and Takala(2002)}]{Freund2001}
Freund, J., Takala, E.-P., 2002. A dynamic model of the forearm including
  fatigue. Journal of Biomechanics, 597--605.

\bibitem[{Garg et~al.(2002)Garg, Hegmann, Schwoerer, and Kapellusch}]{Garg2002}
Garg, A., Hegmann, K., Schwoerer, B., Kapellusch, J., 2002. The effect of
  maximum voluntary contraction on endurance times for the shoulder girdle.
  Industrial Ergonomics, 103--113.

\bibitem[{Giat et~al.(1993)Giat, Mizrahi, and Levy}]{Giat1993}
Giat, Y., Mizrahi, J., Levy, M., July 1993. A musculotendon model of the
  fatigue profiles of paralyzed quadriceps muscle under fes. IEEE Transaction
  on Biomechanical Engineering 40, 664--674.

\bibitem[{HSE(2005)}]{HSE2005}
HSE, 2005. Self-reported work-related illness in 2004/05. Tech. rep., Health,
  Safety and Executive, \url{http://www.hse.gov.uk/statistics/swi/tables/
  0405/ulnind1.htm}.

\bibitem[{Jayaram et~al.(2006)Jayaram, Jayaram, Shaikh, Kim, and
  Palmer}]{Uma2006}
Jayaram, U., Jayaram, S., Shaikh, I., Kim, Y., Palmer, C., 2006. Introducing
  quantitative analysis methods into virtual environments for real-time and
  continuous ergonomic evaluations. Computers In Industry, 283--296.

\bibitem[{J.Russell et~al.(2007)J.Russell, Lori~Winne, E.Camp, and
  W.Johnson}]{Steven2007}
J.Russell, S., Lori~Winne, u., E.Camp, J., W.Johnson, P., 2007. Comparing the
  results of five lifting analysis tools. Applied Ergonomics, 91--97.

\bibitem[{Komura et~al.(1999)Komura, Shinagawa, and L.Kunii}]{KOMURA1999}
Komura, T., Shinagawa, Y., L.Kunii, T., 1999. Calculation and visualization of
  the dynamic ability of the human body. Visualization and Computer Animation,
  57--78.

\bibitem[{Komura et~al.(2000)Komura, Shinagawa, and L.Kunii}]{KOMURA2000}
Komura, T., Shinagawa, Y., L.Kunii, T., 2000. Creating and retargetting motion
  by the musculoskeletal human. The Visual Computer, 254--270.

\bibitem[{Lamkull et~al.(2007)Lamkull, Hanson, and Ortengren}]{Dan2007}
Lamkull, D., Hanson, L., Ortengren, R., 2007. The influence of virtual human
  model appearance on visual ergonomics posture evaluation. Applied Ergonomics,
  713--722.

\bibitem[{Li and Buckle(1999)}]{GUANGYANLI1999}
Li, G., Buckle, P., 1999. Current techniques for assessing physical exposure to
  work-related musculoskeletal risks, with emphasis on posture-based methods.
  Ergonomics 42, 674--695.

\bibitem[{Liu et~al.(2002)Liu, Brown, and Yue}]{Jing2002}
Liu, J.~Z., Brown, R.~W., Yue, G.~H., 2002. A dynamical model of muscle
  activation, fatigue, and recovery. Biophysical Journal, 2344--2359.

\bibitem[{Maier and Ross-Mota(2000)}]{Oregon2000}
Maier, M., Ross-Mota, J., 2000. Work-related musculoskeletal disorders.
  \url{http://www.cbs.state.or.us/external/imd/rasums/ resalert/msd.html}.

\bibitem[{Rodr\'{\i}guez et~al.(2002)Rodr\'{\i}guez, Boulic, and
  D.Meziat}]{Rodriguez2002}
Rodr\'{\i}guez, I., Boulic, R., D.Meziat, 2002. A joint-level model of fatigue
  for the postural control of virtual humans. In: $5^{th}$ Conference of Human
  and Computer. Tokyo.

\bibitem[{Rodr\'{\i}guez et~al.(2003{\natexlab{a}})Rodr\'{\i}guez, Boulic, and
  Meziat}]{Rodriguez20032}
Rodr\'{\i}guez, I., Boulic, R., Meziat, D., Jan.21-22 2003{\natexlab{a}}. A
  model to assess fatigue at joint-level using the half-joint concept. In:
  Visualization and Data Analysis. Santa Clara, CA USA.

\bibitem[{Rodr\'{\i}guez et~al.(2003{\natexlab{b}})Rodr\'{\i}guez, Boulic, and
  Meziat}]{Rodriguez20031}
Rodr\'{\i}guez, I., Boulic, R., Meziat, D., Jan.21-22 2003{\natexlab{b}}.
  Visualizing human fatigue at joint level with the half-joint pair concept.
  In: Visualization and Data Analysis. Santa Clara, CA USA.

\bibitem[{Schaub et~al.(1997)Schaub, Landau, Menges, and Grossmann}]{Shaub1997}
Schaub, K., Landau, K., Menges, R., Grossmann, K., 1997. A computer-aided tool
  for ergonomic workplace design and preventive health care. Human Factors and
  Ergonomics in Manufacturing, 269--304.

\bibitem[{Stanton et~al.(2004)Stanton, Hedge, Brookhuis, Salas, and
  Hendrick}]{Neville2004}
Stanton, N., Hedge, A., Brookhuis, K., Salas, E., Hendrick, H., 2004. Handbook
  of Human Factors and Ergonomics Methods. CRC PRESS.

\bibitem[{{State~of~Washington~Department~of~Labor}(2005)}]{Ergweb}
{State~of~Washington~Department~of~Labor}, 2005. Fitting the job to the worker:
  An ergonomics program guideline. Tech. rep., ERGWEB,
  \url{http://www.ergoweb.com/resources/reference/ guidelines/fittingjob.cfm}.

\bibitem[{Vignes(2004)}]{Vignes2004}
Vignes, R.~M., May 2004. Modeling muscle fatigue in digital humans. Master's
  thesis, Graduate College of The University of Iowa.

\bibitem[{Vollestad(1997)}]{Vollestad1997}
Vollestad, N.~K., 1997. Measurement of human muscle fatigue. Journal of
  Neuroscience Methods, 219--227.

\bibitem[{{VSR~Research~Group}(2004)}]{VSR2004}
{VSR~Research~Group}, October 2004. Technical report for project virtual
  soldier research. Tech. rep., Center for Computer-Aided Design, The
  University of IOWA.

\bibitem[{Wang et~al.(2006)Wang, Zhang, Bennis, and Chablat}]{WANG2006}
Wang, Y., Zhang, W., Bennis, F., Chablat, D., May 2006. An integrated
  simulation system for human factors study. In: The Institute of Industrial
  Engineers Annual Conference. Orlando, Florida.

\bibitem[{Wexler et~al.(1997)Wexler, Ding, and Binder-Macleod}]{Wexler1997}
Wexler, A.~S., Ding, J., Binder-Macleod, S.~A., May 1997. A mathematical model
  that predicts skeletal muscle force. IEEE Transactions On Biomedical
  Engineering.

\end{thebibliography}

\end{document}